# Auto Spell Suggestion for High Quality Speech Synthesis in Hindi

Shikha Kabra
Department of Computer Science and Engineering
Banasthali Vidyapeeth
Jaipur, Rajasthan

Ritika Agarwal
Department of Computer Science and Engineering
Banasthali Vidyapeeth
Jaipur, Rajasthan

## ABSTRACT
The goal of Text-to-Speech (TTS) synthesis in a particular language is to convert arbitrary input text to intelligible and natural sounding speech. However, for a particular language like Hindi, which is a highly confusing language (due to very close spellings), it is not an easy task to identify errors/mistakes in input text and an incorrect text degrade the quality of output speech hence this paper is a contribution to the development of high quality speech synthesis with the involvement of Spellchecker which generates spell suggestions for misspelled words automatically. Involvement of spellchecker would increase the efficiency of speech synthesis by providing spell suggestions for incorrect input text. Furthermore, we have provided the comparative study for evaluating the resultant effect on to phonetic text by adding spellchecker on to input text.

## Keywords
Grapheme; Phoneme; Speech Synthesis; Spellchecker.

## 1. INTRODUCTION
Synthesis system for a particular natural language is the technology for translating or converting a given typed or stored text input into its equivalent spoken waveform format. It is used to translate written information into aural information where it is more convenient especially for mobile applications such as voice enabled mail and messages and also used to assist vision impaired persons so that content of display screen automatically read aloud to a blind user.

People in India commonly use Hindi language for the communication since it is the most popular and commonly used language of all other local Indian languages so it requires representation over Web as an expressive language. Unicode has enabled to read and write over the Web but since it is a highly confusing language due to presence of close spellings (like दीन दिन) Hindi text processing is a major issue which should be given considerable attention. This paper draws attention toward pre-processing of input text by adding Hindi spellchecker on to current TTS system so that text processing could become easier and accurate. Quality of speech synthesis is highly dependent on to quality of input text[1] hence to increase the quality of input text , pre-processing is done by adding spellchecker on to the existing system. The entire process of speech synthesis is divided into three broad parts (figure 1).

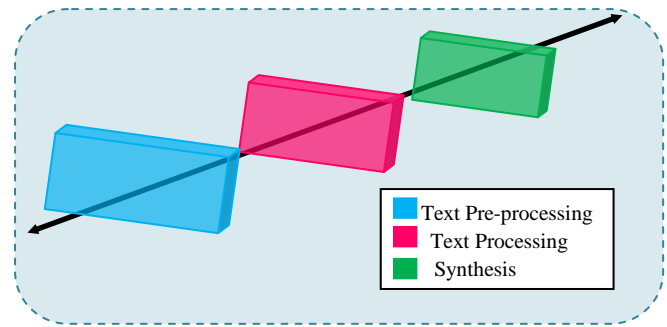

**Fig 1: Process of Text To Speech Synthesis**

First is Text Pre-processing which include correction of manually entered or web extracted text by using Intelligent spellchecker that corrects mistakes/errors with high degree of accuracy by predicting list of most appropriate word to improve written text. After this, improved written text or corrected text is further passed to Text Processing part where it is processed by module (such as normalization, grapheme to phoneme conversion) and converted to phonetic text. Further Phonetic text is passed to Synthesis part where it is converted into its equivalent spoken waveform called output speech. Hence the implementation of above process leads to provide accurate and high quality output speech.

The layout of this paper is as follows. In first section we introduced about the need of spellchecker for a Hindi TTS system. In second section we elaborate the methodology used to convert a written text into speech waveform [2]. In section third we provide comparative study to show the effect on phonetic text by adding spellchecker on to input text. In last section we conclude the paper.

## 2. NEED OF HINDI SPELLCHECKER
Hindi is the most common and popular language of all other Indian languages hence it is more beneficial for Indians to develop TTS system in Hindi language. To develop a Hindi TTS, it is required to enter text in Hindi language which is a highly confusing language, people belongs to different literacy level or having less knowledge of Hindi may not enter the Hindi text exactly correct or may not easily identify errors [3]. Hence Spellchecker is a solution for making input text correct. It is a program that identifies and corrects misspelled word in given input text. Various sources of Hindi input text are:

- Manually entered text
- Online Hindi newspaper, e-mail
- Online Hindi books and journals
- Hindi Websites

With the reference of above sources, text have to be manually entered or extracted from web and may have error in it .Data having error or incorrect data is further processed and used for synthesis which degrade the quality of output speech Types of spelling error could occur due to lack of Hindi spelling knowledge or while typing in Hindi language are [4] real word and non word error. Real word error occurs when misspelled word is also present in language model (collection of words in a particular language) but the word is not fit in context of the sentence and these errors are hard to handle for example:

वह उस और जाता है  (incorrect)

वह उस ओर जाता है  (correct)

मे यहां का राजा हूं  (incorrect)

मैं यहां का राजा हूं   (correct)

Non word error occurs due to wrong key press or lack of spelling knowledge of correct word and mistake arise when word is not found in language model. For example:

अराधना  (incorrect)

आराधना (correct)

धियान  (incorrect)

ध्यान    (correct)

To remove such errors Spellchecker is a solution which enhances the quality of text to be used for speech synthesis.

## 3. PROPOSED APPROACH
In this section we have introduced the approach we have followed for developing high quality speech. The entire process of synthesizing speech is divided into three broad parts (figure 2):

1) Text Pre-processing
2) Text Processing
3) Synthesis

The first task of any text-to-speech (TTS) system is the Pre-Processing of text which includes input text (entered manually or scanned through any online source) and spellchecker (a program corrects misspelled words). Next one is the text processing which includes Text Standardization and Normalization and Grapheme to Phoneme Conversion and the last one is Synthesis which includes waveform generation.

### 3.1  Text Input
Text that a user want to convert into speech signal called text input .For Hindi TTS ,text input should be in Hindi which can be entered manually or extracted from online source but these sources may have error in it. For example:

Text input ⇐ 400 यूनिट तक बजिली इस्तमाल करने वाले लोगो को यू.पी. मे फायदा

Above sentence is taken from online Hindi newspaper where बजिली and इस्तमाल are misspelled words and processing of misspelled text leads to provide incorrect phonetic information which decreases the  naturality of synthetic speech hence to overcome this problem spellchecker is a flexible solution.

### 3.2  Spellchecker
A Spellchecker is a program to locate misspelled word and notify user about the misspellings. Depending on the spellchecker, the feature may either auto correct the word or allow the user to select from list of predicted suggestions of misspelled word. Approach we have used for implementing spellchecker is Dictionary based where we compared every word with list of thousands of properly spelled words to determine most approximated words of misspelled word. For example:

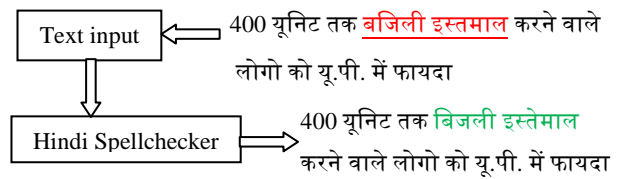

In above example, Hindi text input has some incorrect words like  बजिली and  इस्तमाल and can be  corrected by using Dictionary based spellchecker. Now corrected text is further processed and synthesised which results in high quality speech.

### 3.3  Text Normalization and Standardization
English is world -wide accepted language hence it is preferred to convert Hindi language into standard language like English using WX notation [5].

| अ a | आ A | इ i | ई I | उ u | ऊ U | ए e | ऍ Ez |
|---|---|---|---|---|---|---|---|
| ऐ E | ऑ oz | ओ o | औ O | अं az | अः h | क ka | ख Ka |
| ग ga | घ Ga | ङ fa | च ca | छ Ca | ज ja | झ Ja | ञ Fa |
| ट ta | ठ Ta | ड da | ढ Da | ण Na | त wa | थ Wa | द xa |
| ध Xa | न na | प pa | फ Pa | ब ba | भ Ba | म ma | य ya |
| र ra | ल la | व va | श Sa | ष Ra | स sa | ह ha |  |

**Fig 3: WX Notation sheet**

Problem occurs during standardizing [6] abbreviations, symbols, numbers etc. in Hindi hence to overcome this problem text normalization is a solution which typically involves identification of non-standard words like number abbreviation, symbols and expansion of non-standard words into standard representation without losing its contextual meaning as:

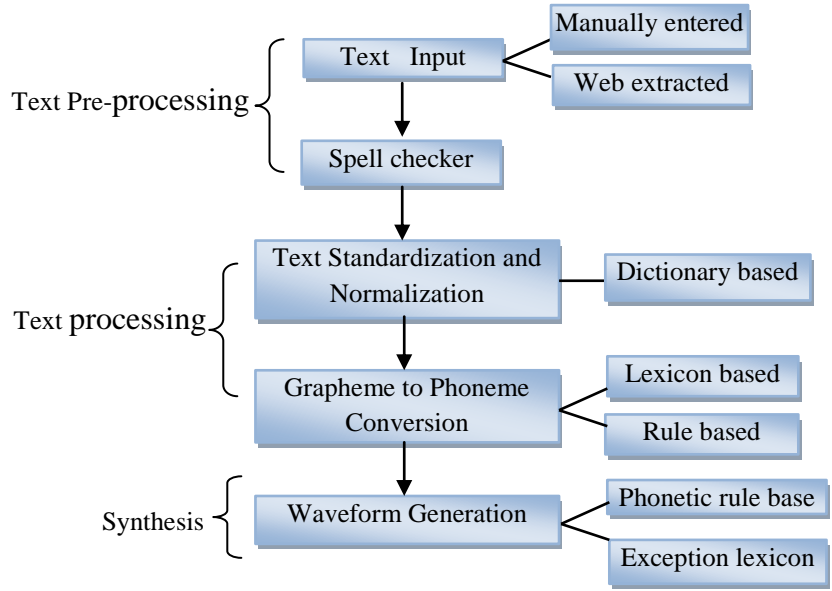

**Fig 2: Approach used for implementing Hindi TTS**

400    cArso

यू.पी.    Uwar praxeS

Since abbreviations and symbols are finite in number, [7] it is easy to normalize them by preparing an exception lexicon but it is not the same case with numbers. It is not feasible to prepare a lexicon for numbers because numbers have no limit and pronounced differently in different context as:

सन्1990-    san UnIso naBe    (year)

1990 किलो-    Ek HazAr no so naBe kIlo    (weight)

Hence this problem can be handled by using rule-based system.

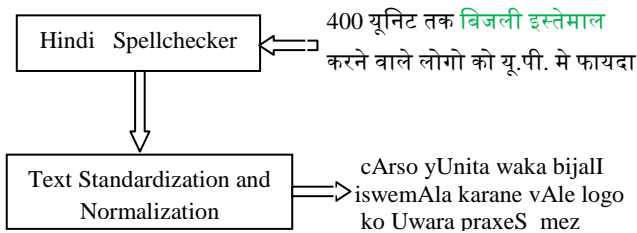

### 3.4 Grapheme to Phoneme Conversion (G2P Conversion)

Grapheme is smallest unit in script similarly phoneme is smallest unit in speech. In [8] Hindi, words are not pronounced in the same way as it is scripted so to provide naturalness in speech it is required to convert grapheme into phoneme to generate phonetic information. But the greatest challenge for high quality speech is misalignment of phonemes at boundaries as:

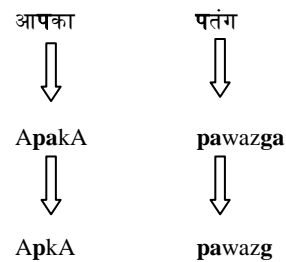

Sound of प varies when place of प changes. In word आपका , प occurs in between ,so pronounced as आधा प (प्  p ) while in पतंग  प occurs at beginning so pronounced as पूरा प ( प  pa)

Basically, this module provides natural pronunciation for a given text which is achieved mainly by manual, rule based or statistical method. Manual approach is most convenient approach but for a lively language like Hindi gets richer day by day , it is impossible to develop a pronunciation dictionary for individual word present. In most of cases, there is one to one correspondence between letter and sound, hence it is easier and more accurate to construct rule based generator rather than statistical generator [9] as:

During G2P conversion if grapheme ends with a sound, like (ta, ma, na...) than (a) gets eliminated from end of grapheme to convert into phoneme. For example:

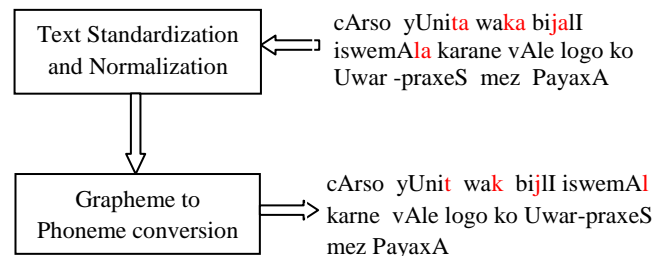

## 3.5 Waveform Generation

Last phase of speech synthesis process is to generate a speech waveform of Phonetic text. We have generated waveform by using open source [10] MARY TTS tool which follows HMM based approach.HMM based approach used for several applications like online fraud detection[11][12],text to speech systems etc. This approach helps to generate natural sounding high quality speech. However, for constructing human-like talking machines, speech synthesis systems are required to have an ability to generate speech with arbitrary speaker's voice characteristics, various speaking styles including native and non-native speaking styles in different languages, varying emphasis and focus, and/or emotional expressions.

## 4. IMPLEMENTATION AND PERFORMANCE EVALUATION

Based on the above approach, Correctness of Speech waveform directly depends onto correctness of Phonetic text hence we have shown comparison between correctness of Phonetic text with/without applying spellchecker onto input text.

आंतकवादी    AnwakavAxI    AnwakvAxI (Antakwadi)

आतंकवादी    AwankavAxI    AwankvAxI (Atankwadi)

From the above observation we analysed that आतंकवादी आंतकवादी are so similar words but dot (.) get misplaced which changes the whole pronunciation from Atankwadi to Antakwadi leads to low quality or incorrect speech.

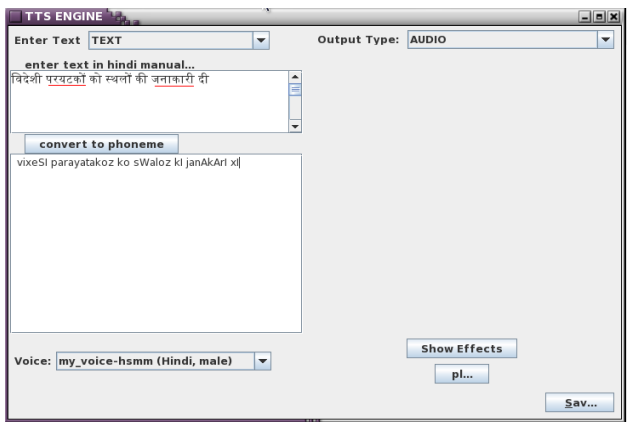

**Fig 4: Snapshot showing incorrect phoneme due to errored input text**

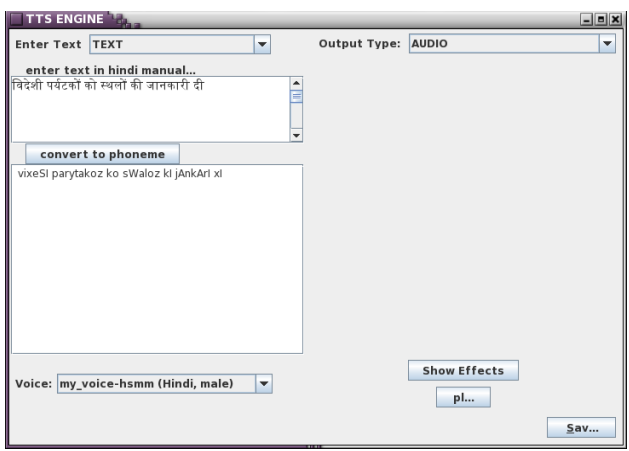

**Fig 5: Snapshot showing correct phoneme after applying spellchecker onto input text**

Figure 4 show that the text to be converted to speech is incorrect and shown by using red underline which leads to wrong phoneme conversion. Figure 5 shows the same text corrected by using spellchecker leads to corrected phoneme. Hence spellchecker is a way to make Phonetic information correct.

## 5. CONCLUSION

In this paper, we presented the development of existing TTS system by adding spellchecker module to it for Hindi language. Since Hindi is a morphologically rich language hence there is a high probability of making mistakes while writing "Mantras" which leads to degrade the quality of speech hence to overcome this problem, Spellchecker is a solution. However, adding spellchecker corrects most of the errored text which leads to increase accuracy of Phonetic text therefore results in high quality/correct speech waveform.

## 6. REFERENCES

[1] Reichel, Uwe D., and Hartmut R. Pfitzinger. "Text preprocessing for speech synthesis." (2006).

[2] Sproat, Richard. "Multilingual text analysis for text-to-speech synthesis."Natural Language Engineering 2.04 (1996): 369-380.

[3] Neha Gupta, Pratistha Mathur "Spell Checking Techniques in NLP: A Survey" International Journal of Advanced Research in Computer Science and Software Engineering, 2012

[4] Sharma, Amit, and Pulkit Jain. "Hindi Spell Checker." (2013).

[5] DATA SHEET FOR WX-NOTATION, http://caltslab.uohyd.ernet.in/wx-notation-pdf/Gujarati-wx-notation.pdf

[6] A. Chauhan, Vineet Chauhan, Surendra P. Singh, Ajay K. Tomar, Himanshu Chauhan" A Text to Speech System for Hindi using English Language" IJCST Vol. 2, Issue 3, September 2011.

[7] Macchi, Marian. "Issues in text-to-speech synthesis." Intelligence and Systems, 1998. Proceedings., IEEE International Joint Symposia on. IEEE, 1998.

[8] S. S. Agrawal, "Synthesizing Hindi speech using Klsyn and Hlsyn for natural sounding," in Workshop on Spoken Language Processing, TIFR, Mumbai, 2003.

[9] Murthy, Hema A., et al. "Building Unit Selection Speech Synthesis in Indian Languages: An Initiative by an Indian Consortium." Proceedings of COCOSDA, Kathmandu, Nepal (2010).

[10] Schröder, Marc, and Jürgen Trouvain. "The German text-to-speech synthesis system MARY: A tool for research, development and teaching." International Journal of Speech Technology 6.4 (2003): 365-377.

[11] Ankit Mundra and Nitin Rakesh. "Online Hybrid Model for Online Fraud Prevention and Detection." Intelligent Computing, Networking, and Informatics. Springer India, 2014. 805-815.

[12] Ankit Mundra & Rakesh, N. (2014, January). Online Hybrid Model for Fraud Prevention (OHM-P): Implementation and Performance Evaluation. In ICT and Critical Infrastructure: Proceedings of the 48th Annual Convention of Computer Society of India-Vol II (pp. 585-592). Springer International Publishing.